\documentclass[twocolumn]{article}
\usepackage{graphicx} 
\usepackage{titling}
\usepackage{amsmath}
\usepackage{graphicx}
\usepackage{placeins} 

\title{Dueling Deep Reinforcement Learning for Financial Time Series}

\author{
  Bruno Giorgio \\
  Independent Researcher \\
  London, United Kingdom \\
}

\date{}

\begin{document}
\maketitle

\begin{abstract} 
Reinforcement learning (RL) has emerged as a powerful paradigm for solving decision-making problems in dynamic environments. In this research, we explore the application of Double DQN (DDQN) and Dueling Network Architectures, to financial trading tasks using historical SP500 index data. Our focus is training agents capable of optimizing trading strategies while accounting for practical constraints such as transaction costs. The study evaluates the model performance across scenarios with and without commissions, highlighting the impact of cost-sensitive environments on reward dynamics. Despite computational limitations and the inherent complexity of financial time series data, the agent successfully learned meaningful trading policies. The findings confirm that RL agents, even when trained on limited datasets, can outperform random strategies by leveraging advanced architectures such as DDQN and Dueling Networks. However, significant challenges persist, particularly with a sub-optimal policy due to the complexity of data source.
\end{abstract}

\section*{Introduction}
Deep Reinforcement Learning (DRL) for financial trading has evolved significantly over the past few years, showcasing a diverse array of methodologies and applications. The rapid change of the increasing amount of data available have revolutionize the finance industry. Classical financial theory has been based on stochastic control for decades as the foundations of finance. These models sometimes oversimplify the mechanism and behaviour of financial markets. On the other hand, models that capture the complexity of market are mathematically and computationally not feasible using the classical tool (Hambly et al., 2022). \\
The complexities of financial markets such as non-linearity, volatility, and high-frequency trading, makes Reinforcement Learning (RL) particularly appealing in this context. Agent acting with an environment might learn to make optimal decision (policy) through repeated experiences in the system applying RL algorithms in areas like order execution, market making and portfolio optimization. \\
Solving a reinforcement learning task means, roughly, finding a policy that achieves the most reward over the long run. There are various reinforcement learning (RL) algorithms, and in this project, we focus on the Double DQN. Traditional Q-learning is known to overestimate Q-values due to the maximization step during value updates. This overestimation issue persists even when using function approximation. Double DQN addresses this problem by decoupling the action selection and action evaluation steps during the target value computation (Van Hasselt et al, 2015). The max operator in the DQN utilizes the same values both to select and evaluate the action $
\max_{a'} Q(s', a'; \theta')$. To avoid this situation and overestimation, we need to decouple the selection from evaluation according to an idea that comes from Van Hasselt about the Double Q-Learning (Van Hasselt, 2010). 
In this process, the performed action is based on a network with weights $\theta$, while the action is evaluated with a second network (target) with weights $\theta'$ considering the next state which can be formally denoted as follows,
$$
Y_t^{\text{DDQN}} = R_{t+1} + \gamma Q\left(s', \arg\max_{a'} Q\left(s', a'; \theta_t\right); \theta_t' \right)
$$
The DDQN algorithm can be seen as an extension of the DQN, with the key feature that it additionally uses the target network to separate the execution and evaluation process of action. 

\section*{Model}
The model focuses on a Dueling DDQN architecture for financial time series trading using two architectures: Feedforward (FFDQN) and Convolutional Neural Network (CNN). Both models are trained on complex financial data, aiming to identify patterns and optimize trading strategies. The study involves testing various hyperparameters, particularly the batch size, to understand its impact on performance. Smaller batch sizes (32-bit) are expected to capture short-term nuances but may introduce noisier gradients, while larger batch sizes (128-bit) are hypothesized to improve stability and generalization, especially in environments with transaction costs. \\
We employ the StockEnv API—a custom Gymnasium environment for stock trading. Our RL system is implemented using the open-source PTAN library, which simplifies code by abstracting actions, experience replay, and environment interaction, as introduced by Maxim Lapan (2018). To formulate an RL problem, we need the Environment's observation, possible Actions, and a Reward system. The raw data consists of nearly 1 million rows of SP500 index data from 2018 to 2019, with five elements per minute: Open, High, Low, Close prices (as percentages of the Open price), and Volume. The observation spans multiple data points, allowing the agent to analyze N consecutive data points to detect patterns and trends over time, aiding informed decision-making. 

\begin{figure*}[htb]
  \centering
  \includegraphics[width=0.9\textwidth]{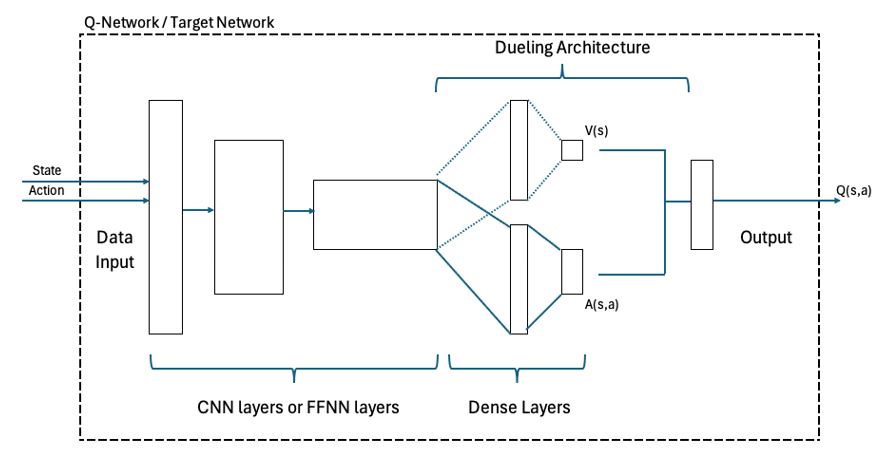}
  \caption{Implementation of Dueling Architecture into the Q-Network/Target Network}
  \label{fig:wide_image}
\end{figure*}

About the reward, the typical method for examining the changes in stock market prices is to look at the so-called returns rather than the actual prices. The reward function is designed based on the agent's actions and trading status at each time step. If the agent chooses to take no action (i.e., no position is opened), the reward at time step t is considered null. When the agent opens a position (buy action), a trading commission is deducted, and a reward is computed based on the market's immediate response. Similarly, when the agent closes a position (sell action), a commission is applied, and the corresponding reward is calculated at time t. Unlike traditional approaches that compute rewards only at the end of an episode, this design evaluates rewards at every time step with an open position. This allows for more frequent feedback, which accelerates convergence and supports more effective learning in dynamic financial environments.

While reinforcement learning (RL) theory does not provide universal convergence guarantees (Sutton \& Barto, 2018), the stability and performance of learning algorithms can be significantly improved through the integration of deep learning techniques, as demonstrated by Mnih et al. (2015). The algorithm implemented in this work is inspired by their Deep Q-Network (DQN) framework. Specifically, the environment—built using OpenAI Gymnasium—generates market states that are fed into the Q-network to determine action-value estimates. The resulting transitions, comprising state, action, reward, and next state, are stored in the replay buffer (see Fig. 4). During training, a mini-batch of transitions is sampled from the replay buffer to compute the target Q-values Q(s, a), using the maximum estimated Q-value of the next state maxQ(s', a'). These target values are then used to calculate the loss, which guides the backpropagation process for updating the Q-network. The agent follows an $\epsilon$-greedy policy, starting fully random ($\epsilon$=1) and gradually shifting to exploitation ($\epsilon$=0.1).
\section*{Dueling Architecture}
The Dueling Architecture (Wang et al., 2016) is divided into two separate parts: Value Function Stream and Advantage Function Stream.

-	Value Function $V(s)$: This stream estimates the state value $V(s)$, which represents the intrinsic value of being in a particular state, irrespective of the action taken.

-	Advantage Function $A(s, a)$: This stream estimates the advantage function $A(s, a)$, which measures the relative importance of taking a specific action in a given state. 

During the process, the input—a batch of states—is passed through the network to initiate both the Value $V(s)$ computation and the Advantage $A(s, a)$ computation streams. The resulting Q-values are calculated using the formula derived from Wang et al. (2016), combining these components as follows:
$$
Q(s, a) = V(s) + \left( A(s, a) - \frac{1}{|\mathcal{A}|} \sum_{a'} A(s, a') \right)
$$
$Q(s, a)$ is the state-action value function. $V(s)$ is the value of being in state. $A(s, a)$ is the advantage of taking action a in state s. $\left| A \right|$  is the cardinality of the action space. The term $\frac{1}{|A|} \sum_{a' \in A} A(s, a') $ is the mean advantage. In fig.1 there is a schema about how has been implemented into Q/Target networks. This separation enhances generalization across actions, improves policy evaluation, and accelerates learning by reducing overestimation bias, especially when combined with techniques like Double DQN.\\

\section*{Training}
The model processes observations over N time steps. Volume data is a challenge due to its inconsistent correlation with price movements. While MACD is a popular momentum indicator, it wasn’t used in RL training since its time-dependence conflicts with the random sampling in experience replay. We trained with batch sizes between 32–128 and included a 1\% commission per trade on the SP 500 Futures. \\
During training, we monitored Reward performance across steps while experimenting with different batch sizes (32 and 128) for FFDQN and CNN models. The training spanned ~8 million steps (~200K episodes on the 1-minute SP500 data). The reward curves below compare performance across batch sizes, showing slight upward trends for FFDQN (batch 128, blue) and CNN (batch 128, yellow), despite typical RL reward oscillations. \\
\begin{figure}[htbp]
  \centering
  \includegraphics[width=\columnwidth]{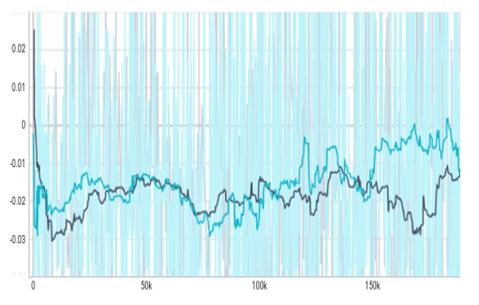}  
  \caption{Reward \% (y-axis) over Training Episodes (x-axis). FFDQN model with Batch size 32 bit (black line) and Batch size 128 (light blue line).}
  \label{fig:one_column}
\end{figure}
\begin{figure}[htbp]
  \centering
  \includegraphics[width=\columnwidth]{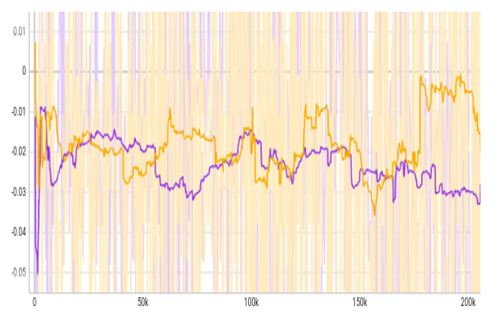}  
  \caption{Reward \% (y-axis) over Training Episodes (x-axis). CNN model with Batch size 32 bit (purple line) and Batch size 128 (yellow line)).}
  \label{fig:two_column}
\end{figure}

Financial markets are inherently stochastic and non-stationary, making it difficult for the agent to learn stable policies. This aligns with observations in the literature, such as those by Jiang et al. (2017) and Zejnullahu et al. (2022), which emphasize the difficulty of applying RL in financial markets due to their unpredictable and dynamic nature.

\section*{Testing}
\begin{figure*}[htbp]
  \centering
  \includegraphics[width=\linewidth]{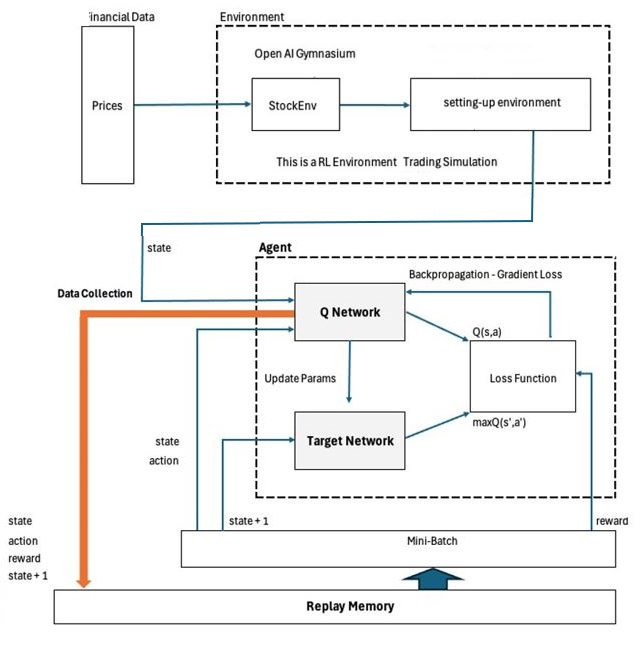}  
  \caption{Logic schema of Training of DDQN architecture.}
  \label{fig:three_column}
\end{figure*}

Testing the FFDQN, the smaller batch size (32) led to noisier gradients and less stable performance, especially with commissions. This suggests potential overfitting or declining rewards due to transaction costs. Below, we compare 32-batch model with and without commissions. Only the no-commission scenario ends with nearly 10\% annual cumulative reward (Fig. 5 \& 6). \\
\begin{figure}[htbp]
  \centering
  \includegraphics[width=8cm, height=7cm]{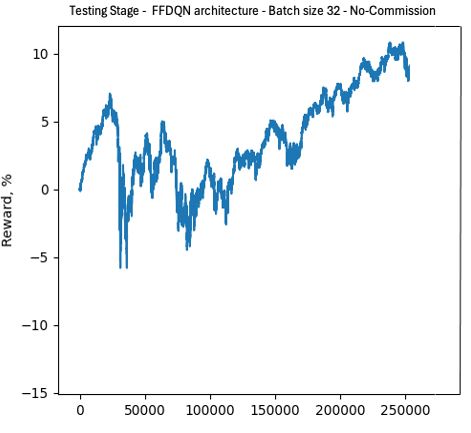}  
  \caption{Testing Cumulative Reward over Training Episodes on FFDQN model batch size 32-bit without commission costs.}
  \label{fig:four_column}
\end{figure}
\begin{figure}[htbp]
  \centering
  \includegraphics[width=8cm, height=7cm]{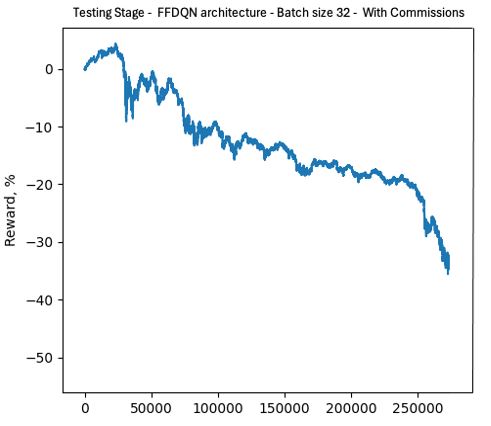}  
  \caption{Testing Cumulative Reward over Training Episodes on FFDQN model batch size 32-bit with commission costs.}
  \label{fig:five_column}
\end{figure}
Changing the hyperparameter to 128-bit Batch size, positive rewards in both cases generate profits.  A batch size of 128 shows slightly better performance, suggesting it benefits model learning with 11\% and 8\% of annual cumulative rewards in no-commission and commission scenarios (Fig. 7 \& 8).
\begin{figure}[htbp]
  \centering
  \includegraphics[width=8cm, height=7cm]{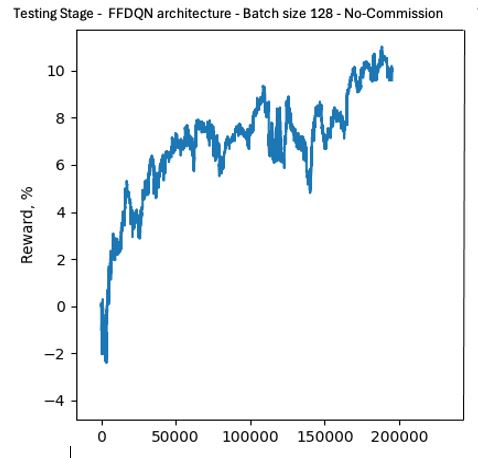}  
  \caption{Testing Cumulative Reward over Training Episodes on FFDQN model batch size 128-bit without commission costs.}
  \label{fig:six_column}
\end{figure}
\begin{figure}[htbp]
  \centering
  \includegraphics[width=8cm, height=7cm]{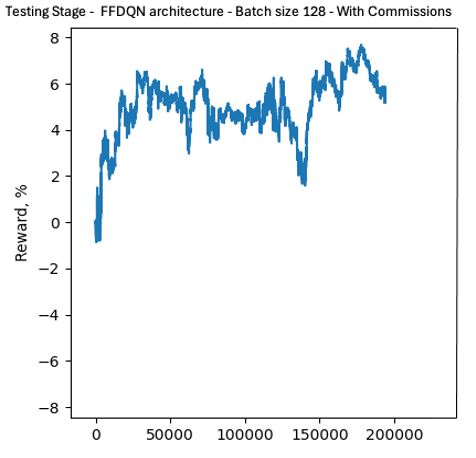}  
  \caption{Testing Cumulative Reward over Training Episodes on FFDQN model batch size 128-bit with commission costs.}
  \label{fig:seven_column}
\end{figure}

The second model we implemented incorporates convolutional layers to extract meaningful features from the input data. Convolutional Neural Networks (CNNs) are particularly effective at capturing temporal dependencies and local patterns in sequential data. While the training procedure closely resembles that of the FFDQN model, the CNN-based architecture provides an enhanced ability to learn both spatial and temporal representations from financial time series. In the 32-bit batch model we have a positive annual reward of 20\% under the no-commission scenario, while incurring a loss when trading costs are included (Fig. 9 \& 10). \\
\begin{figure}[htbp]
  \centering
  \includegraphics[width=8cm, height=7cm]{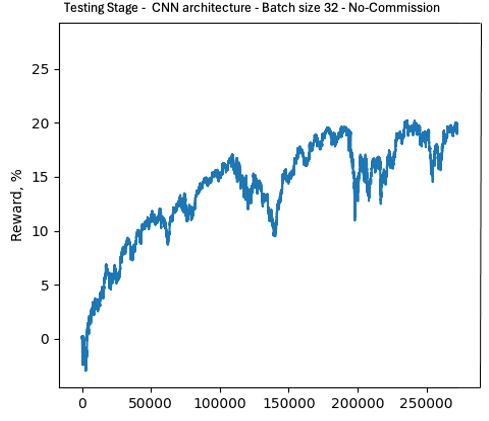}  
  \caption{Testing Cumulative Reward over Training Episodes on CNN model batch size 32-bit without commission costs.}
  \label{fig:one_column_2}
\end{figure}
\begin{figure}[htbp]
  \centering
  \includegraphics[width=8cm, height=7cm]{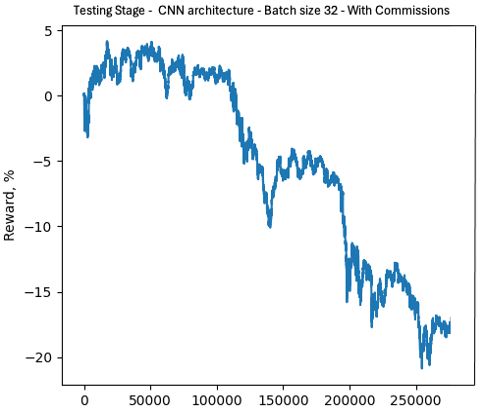}  
  \caption{Testing Cumulative Reward over Training Episodes on CNN model batch size 32-bit without commission costs.}
  \label{fig:heigt_column}
\end{figure}
This observation reinforces the inherently noisy behavior associated with the 32-sample batch size. As noted by Keskar et al. (2017), smaller batch sizes often struggle to generalize effectively, especially in complex environments that incorporate transaction costs or penalty mechanisms. In contrast, larger batch sizes (128-bit) are more stable and accurate gradient estimates due to their ability to average over a greater number of data points. \\
\begin{figure}[htbp]
  \centering
  \includegraphics[width=8cm, height=7cm]{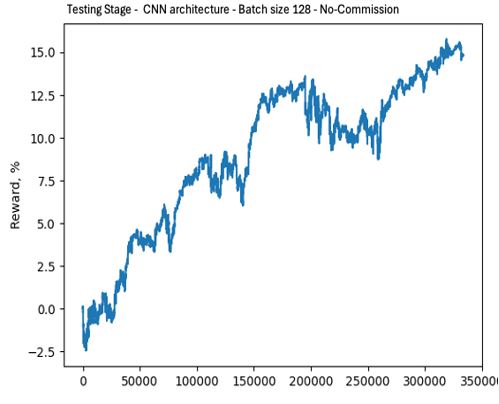}  
  \caption{Testing Cumulative Reward over Training Episodes on CNN model batch size 128-bit without commission costs.}
  \label{fig:nine_column}
\end{figure}
\begin{figure}[htbp]
  \centering
  \includegraphics[width=8cm, height=7cm]{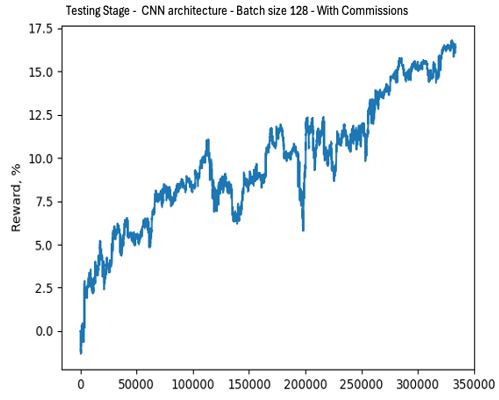}  
  \caption{Testing Cumulative Reward over Training Episodes on CNN model batch size 128-bit without commission costs.}
  \label{fig:ten_column}
\end{figure}
This is evident in the case of the 128-bit batch size, which consistently achieves reward levels exceeding 15\% across both scenarios (Fig. 11 \& 12). \\

About the commission behaviour, where negative rewards dominate in scenarios with commission costs is well-documented in finance RL. Our results are consistent with the observation. Agents often engage in excessive trading, especially during exploration phases, leading to the accumulation of transaction costs and negative net rewards. In fact, Zhang et al. (2021) discuss how RL agents tend to overfit to short-term opportunities during training. When commissions are introduced, these short-term trades incur cumulative costs, resulting in negative overall rewards. One possible solution is based on the regularization techniques where we penalize frequent trades by adding a regularization term to the reward function (Zhang et al., 2021). \\

\FloatBarrier

\section*{Conclusion}
This essay shows that DDQN with Dueling Architecture is a promising RL approach for financial trading, achieving positive returns—especially in no-commission settings—and adapting well to transaction costs with larger batch sizes.
\begin{figure}[htbp]
  \centering
  \includegraphics[width=\columnwidth]{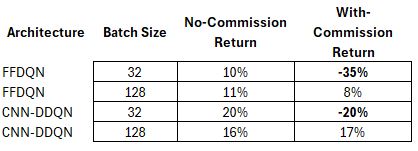}  
  \caption{Annual Return (\%) after one year of real-world SP500 trading}
  \label{fig:eleven_column}
\end{figure}
First, the FFDQN 32-bit batch model exhibited a strong sensitivity to transaction costs: it achieved a +10\% return without commission, but performance deteriorated sharply to –35\% with commission.
In contrast, the FFDQN 128-bit model performed more consistently, with +11\% return without commission and +8\% with commission, suggesting better generalization. Increasing the batch size appears to have a regularizing effect, likely by reducing variance in Q-value estimates during training (Henderson et al., 2018). \\
For CNN-based models, the 32-bit batch version again shows signs of strong sensitivity, returning +20\% with no commission but dropping to –20\% with commissions.
The CNN 128-bit batch version, however, was the most robust, achieving +16\% without commission and +17\% with commission—even improving under realistic trading constraints. This robustness may suggest that the larger batch size helped the model learn more stable and efficient policies. \\

The results suggest that the 32-bit model likely overfitted to short-term, noisier market trends, failing to generalize effectively under realistic trading conditions where transaction costs are present. Overfitting in reinforcement learning (RL) differs from the traditional supervised learning context. In RL the objective is to learn an optimal policy by maximizing expected cumulative rewards in a non-stationary and often partially observable environment (Sutton \& Barto, 2018). While RL models do not "overfit" in the conventional sense, they can still converge to sub-optimal policies due to poor exploration, limited experience replay, or inadequate generalization to unseen states (Zhang et al., 2018). \\

The financial stock market is generally harder to model compared to Atari 2600 games due to its complex state space, stochastic and non-stationary dynamics, sparse and noisy rewards, and higher computational demands. These differences underscore why RL models for financial markets require advanced techniques and computational resources. Mnih, Silver et al. (2015) trained their RL models on Atari games for approximately 7 to 10 days per game, with over 40 games in total. In comparison, financial RL models often deal with even larger datasets. My model was trained on  1-min range of index data and in almost one day training. Training an accurate trading policy would require extensive data from 1900 onward and significant computational resources. \\

\section*{References}
\noindent
\hangindent=10mm
\hangafter=1
Hambly, B., Xu, R., \& Yang, H. (2022) Recent Advances in Reinforcement Learning Finance. \\

\noindent
\hangindent=10mm
\hangafter=1
Henderson, P., Islam, R., Bachman, P., Pineau, J., Precup, D., \& Meger, D. (2018). Deep Reinforcement Learning that Matters. In Proceedings of the AAAI Conference on Artificial Intelligence, 32(1). \\

\noindent
\hangindent=10mm
\hangafter=1
Jiang, Z., Xu, D., \& Liang, J. (2017). A deep reinforcement learning framework for the financial portfolio management problem. arXiv preprint arXiv:1706.10059. \\

\noindent
\hangindent=10mm
\hangafter=1
Keskar, N. S., et al. (2017). On Large-Batch Training for Deep Learning: Generalization Gap and Sharp Minima. arXiv preprint arXiv:1609.04836. \\

\noindent
\hangindent=10mm
\hangafter=1
Lapan, M. (2018). Deep Reinforcement Learning Hands-On: Apply modern RL methods to practical problems of chatbots, games, robotics, and stock trading with PyTorch. Birmingham: Packt Publishing. \\

\noindent
\hangindent=10mm
\hangafter=1
Mnih, V., Kavukcuoglu, K., Silver, D., Rusu, A.A., Veness, J., Bellemare, M.G., Graves, A., Riedmiller, M., Fidjeland, A.K., Ostrovski, G., Petersen, S., Beattie, C., Sadik, A., Antonoglou, I., King, H., Kumaran, D., Wierstra, D., Legg, S. and Hassabis, D. (2015) Human-level control through deep reinforcement learning. Nature, 518(7540), pp.529- 533 \\

\noindent
\hangindent=10mm
\hangafter=1
Sutton, R. \& Barto, A. (2018) Reinforcement Learning. An Introduction. MIT Press. Cambridge \\

\noindent
\hangindent=10mm
\hangafter=1
Van Hasselt, H., Guez, A., \& Silver, D. (2015). Deep reinforcement learning with double Q- learning. Proceedings of the Thirtieth AAAI Conference on Artificial Intelligence (AAAI-16), Phoenix, Arizona, USA. \\

\noindent
\hangindent=10mm
\hangafter=1
Van Hasselt, H. (2010) Double Q-Learning. Advances in Neural Information Processing Systems. 23:2613-2621 \\

\noindent
\hangindent=10mm
\hangafter=1
Zejnullahu, F., Moser, M., \& Osterrieder, J., (2022) Applications of Reinforcement Learning in Finance - Trading with a Double Deep Q-Network. \\

\noindent
\hangindent=10mm
\hangafter=1
Wang, Z., Schaul, T., Hessel, M., Van Hasselt, H., Lanctot, M. \& De Freitas, N. (2016). Dueling Network Architectures for Deep Reinforcement Learning. arXiv preprint arXiv:1511.06581. Available at: https://arxiv.org/abs/1511.06581 [Accessed Date 10 November 2024]. \\

\noindent
\hangindent=10mm
\hangafter=1
Zhang, Y., Yang, X. \& Li, D. (2021). Enhancing trading strategies using reinforcement learning combined with supervised learning and genetic algorithms. IEEE Transactions on Computational Intelligence and AI in Games. \\

\noindent
\hangindent=10mm
\hangafter=1
Zhang, C., Vinyals, O., Munos, R., \& Bengio, S. (2018). A study on overfitting in deep reinforcement learning. arXiv preprint arXiv:1804.06893

\end{document}